\pdfoutput=1
\documentclass[11pt]{article}
\usepackage[preprint]{acl}
\usepackage{times}
\usepackage{latexsym}
\usepackage[T1]{fontenc}
\usepackage[utf8]{inputenc}
\usepackage{microtype}
\usepackage{inconsolata}
\usepackage{booktabs}
\usepackage{amsmath}
\usepackage{graphicx}
\usepackage{xcolor}

\title{Scoring Both Directions: LLMs realize the MRS they cannot reliably parse}

\author{Soham Dan \\
  Scale AI \\
  \texttt{soham.dan@scale.com}}

\begin{document}
\maketitle

\begin{abstract}
The English Resource Grammar (ERG) is a hand-written computational grammar of English. Given a sentence, its processor, ACE, produces a formal meaning representation called Minimal Recursion Semantics (MRS): a graph of the sentence's predicates and their arguments. The grammar is bidirectional and can also turn an MRS back into an English sentence. \citet{hajdik2019} used the ERG's treebank to build a benchmark for that generation task, MRS to text, and trained sequence-to-sequence models to solve it. The parsing task, text to MRS, can be tested on the same sentences. We reconstruct their 10K-sentence test split, and score two large language models, Claude Sonnet~4.5 and Claude Opus~5, in both directions against their trained systems and against ACE, with no task-specific training. Given an MRS and three examples, Opus writes the sentence at 76.3 BLEU, ten points above their system trained on 72k pairs (66.1 BLEU), and comparable to their system trained on a million extra pairs (77.2 BLEU). Sonnet scores 65.7 BLEU, and letting it choose among ACE's own candidate sentences lifts it to 69.6, while a pooled judge that keeps Opus's own sentence among the candidates adds 0.6 points (77.0 BLEU). In the parsing direction, however, the models fall far behind ACE: asked for the MRS of the same sentences, they reach 57.2 (Sonnet) and 65.5 (Opus) F$_1$ on the graph's predicates and arguments against 91.0 for ACE, and exact-match the gold on about 1\% of sentences. We characterize the failure modes for the parsing tasks, and conclude that a generation score alone does not show that models understand formal semantic representations.
\end{abstract}

\begin{figure}[t]
\centering
\includegraphics[width=\columnwidth]{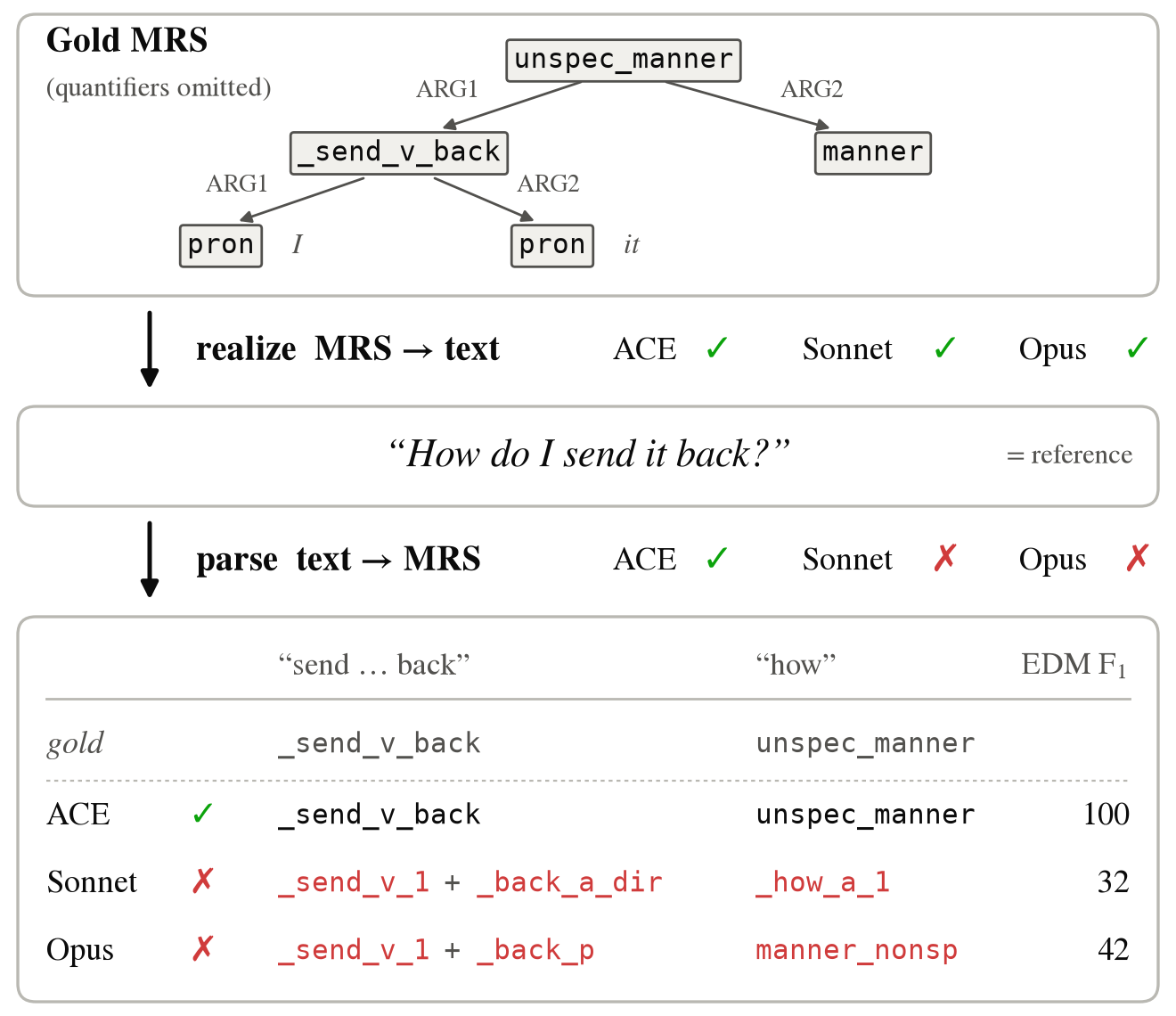}
\caption{An example sentence where all three systems realize the reference exactly and only ACE parses it exactly. Given the gold MRS (linearized for the LLMs), ACE, Sonnet and Opus all write the reference sentence. Given the sentence, only ACE recovers the MRS. Both LLMs split the particle verb \emph{send back}, which the ERG treats as one predicate, and use predicates for \emph{how} that the ERG does not have. Red marks predicates that differ from gold. The last column is the spans-off EDM F$_1$.}
\label{fig:headline}
\end{figure}

\section{Introduction}

The English Resource Grammar (ERG; \citealp{flickinger2000erg}) is a hand-written, broad-coverage grammar of English. Its processor, ACE \citep{crysmann2012}, parses a sentence into Minimal Recursion Semantics (MRS; \citealp{copestake2005mrs}), a graph whose nodes are the sentence's predicates and whose edges are their argument roles, with quantifier scope left underspecified, and runs backwards to generate from a given MRS the sentences it represents. The Redwoods treebank \citep{oepen2004redwoods} pairs sentences from newswire, Wikipedia, tourism prose, fiction and dialogue with their hand-selected gold MRS (about ten thousand in its test partition). \citet{hajdik2019} turned this treebank into a benchmark for the generation direction, MRS to text: each gold MRS linearized (as Dependency MRS, \citealp{copestake2009}, in PENMAN form), trained a sequence-to-sequence model on 72k pairs, using BLEU to score against the original sentence. Their systems reach 66.11 BLEU on the 10K test split, and 77.17 with about a million extra silver pairs parsed automatically from newswire. Because the grammar is bidirectional, the same split poses the reverse task, text to MRS, on the same sentences, with ACE as a reference in both directions. A prompted language model needs no task-specific training for either, so the same system can be scored both ways on identical items. We use Claude Sonnet~4.5 and Claude Opus~5 with three exemplars each, and ask three research questions.

\paragraph{Does a prompted model reach the trained realizer?} Yes, and Opus beats it (Table~\ref{tab:main}): Opus~5 scores 76.34 BLEU, 10 points above the 72k-pair trained seq2seq system, and on par with the million-pair silver-data trained system. Sonnet~4.5 obtains 65.65 BLEU matching the trained 72k-pair system. These public sentences predate both models' training cutoffs and we run a substitution control to test memorization which shows relatively small memorization effects(\S\ref{sec:real}). Parsed back through the grammar, the models' sentences recover the input MRS at 85.4 (Sonnet) and 89.8 (Opus) spans-off EDM F$_1$ against 91.0 for the gold sentences themselves, so high BLEU here reflects preserved meaning, not surface overlap alone.

\paragraph{Does the grammar help?} It helps the weaker model, Sonnet: ACE generates candidates for 79\% of the items, its top-10 best list holds realizations its ranker does not surface, and Sonnet as a fluency judge over that list (using Sonnet's own sentence where ACE abstains) reaches 69.64 BLEU on the test split, four points over Sonnet alone (Table~\ref{tab:main}). For Opus, we use a pooled judge that keeps Opus's own sentence among the candidates and this shows a gain of about $0.6$ BLEU, which shows that the grammar helps both models with diminishing returns for the stronger model. 

\paragraph{Can the models parse back the MRS from the sentence?} Not reliably. When we ask for the MRS of the same sentences, Sonnet reaches only 57.2 span-free EDM F$_1$ and Opus 65.5 compared to 91.0 for ACE's parser, matching the gold analysis exactly on 0.6\% and 1.4\% of items against 50.5\% for ACE (Table~\ref{tab:parse}). The outputs are structure-formatted correctly and almost always decode as MRS but what fails is the grammar's conventions, the predicate inventory and the argument frames (\S\ref{sec:parse}).

Limits of LLMs as AMR analysts have been reported \citep{ettinger2023} and the producing/understanding gap mentioned in \citet{west2024}, ERG lets us measure it on the same benchmark with a grammar as the reference both ways. 

\section{Setup}
\label{sec:setup}

\paragraph{Benchmark.} We take the ERG-1214 gold profiles and run \citeauthor{hajdik2019}'s released conversion unchanged, so inputs and references match their files  exactly yielding 10,201 test items over 37 Redwoods profiles. 

\paragraph{Three tasks.} Q1, \emph{realization}: linearized MRS $\rightarrow$ sentence over the full test split. Q2, \emph{reranking}: Sonnet, as a fluency judge, chooses among ACE's 10-best sentences, the model's own prediction filling in where ACE produces nothing (\emph{hybrid}) and a \emph{pooled} variant that adds the model's own sentence to the list. Q3, \emph{analysis}: sentence $\rightarrow$ SimpleMRS on a fixed-seed 1{,}000-item sample stratified on the newswire data (whose gold MRS passes through the evaluator unchanged) are scored.

\paragraph{Models and prompts.} Claude Sonnet~4.5 (\texttt{20250929}) and Claude Opus~5 with three fixed exemplars from outside the scored items. Sonnet ran with greedy decoding and a 3{,}000-token budget, and Opus~5 runs with default thinking. 

\paragraph{The grammar.} ACE~0.9.31 with ERG~1214, as realizer ($n$-best 10, 10~s per item) and parser (top-1). ACE is in-sample: the gold MRSs are ERG-1214 analyses, and its ranking model was trained on Redwoods, including these profiles. 

\paragraph{Metrics.} Realization is scored by SacreBLEU (exponential smoothing) \cite{post2018} against the original sentence after their post-processing. Parsing is scored on the EDS reduction of each MRS \citep{oepen2006eds} with three metrics (App.~\ref{app:spec}). EDM F$_1$ \citep{dridan2011} matches predicates and argument triples, either anchored to their character spans or with spans removed so that nodes are keyed by predicate name alone. Exact match requires the predicted and gold graphs to be isomorphic once spans are ignored. Smatch \citep{cai2013} matches triples under the node alignment that maximizes agreement, so it is invariant to variable names. 

\begin{table}[t]
\centering\scriptsize\setlength{\tabcolsep}{18pt}
\begin{tabular}{lr}
\toprule
System & BLEU \\
\midrule
\multicolumn{2}{l}{\emph{trained systems \citep{hajdik2019}}} \\
\quad gold-only (72k pairs)                &     66.11 \\
\quad gold+silver ($\sim$1M pairs)         &        77.17 \\
\midrule
\multicolumn{2}{l}{\emph{grammar}} \\
\quad ACE 1214, generable (79.2\%)        & 61.81  \\
\midrule
\multicolumn{2}{l}{\emph{prompted LLMs with three exemplars}} \\
\quad Sonnet 4.5                           & 65.65  \\
\quad Opus 5                               & \textbf{76.34} \\
\midrule
\multicolumn{2}{l}{\emph{model + grammar (Q2)}} \\
\quad Sonnet + ACE, judge hybrid           & 69.64  \\
\quad Opus + ACE, pooled judge             & 76.95  \\
\bottomrule
\end{tabular}
\caption{Q1 realization on \citeauthor{hajdik2019}'s split and the Q2 hybrids (Sonnet as judge throughout: its pick where ACE generates, the model's prediction elsewhere). }
\label{tab:main}
\end{table}

\section{Realization}
\label{sec:real}

\paragraph{Q1: LLMs against the trained realizer.} Table~\ref{tab:main} shows both LLMs with three prompted exemplars, evaluated on the same linearized MRS test set as the trained systems from \citet{hajdik2019}. Sonnet~4.5 matches the trained 72k-pair (gold-only) system and Opus~5 is 8-10 points above it essentially tied with the million-pair (gold+silver) trained system. We also compare this to ACE which realizes 79.2\% of the split at 61.81 BLEU.

\paragraph{Is high BLEU faithful?} BLEU measures overlap with the reference, not preservation of the input, so we parse each system's sentence back with ACE and score the parse against the input MRS: its parse of the gold sentence itself scores 91.0 spans-off EDM F$_1$ on the 1K items (92.6 on the 762 ACE generates for), exact on 50.5\%. Sonnet's sentences score 85.4 (exact 28.6\%; 96.1\% parse), Opus's 89.8 (38.9\%; 96.7\%); ACE's own realization and the judge's pick score 94.5 and 94.3 on the 762. Opus's sentences sit within 1.2 points of the gold sentence's own reparse, Sonnet's within 5.6: the realizations largely preserve the input's predicates and arguments (App.~\ref{app:spec}).

\begin{table}[t]
\centering\footnotesize\setlength{\tabcolsep}{3pt}
\begin{tabular}{lr}
\toprule
items ACE generates for ($n{=}7{,}865$) & BLEU \\
\midrule
ACE 1-best (its own order)                            & 61.82 \\
shortest candidate                                    & 69.01 \\
Sonnet as fluency judge over ACE's $n$-best           & 71.25 \\
reference selector over ACE's $n$-best$^{c}$          & 74.84 \\
\midrule
Sonnet 4.5 alone                                      & 67.02 \\
Opus 5 alone                                          & 77.50 \\
reference selector, ACE's $n$-best $\cup$ Opus & 83.07 \\
Sonnet judge, ACE's $n$-best $\cup$ Opus              & \textbf{78.36} \\
\bottomrule
\end{tabular}
\caption{Q2, where ACE generates. $^{c}$Per item the highest sentence-BLEU candidate which is a ceiling. Sonnet scores 62.56 and Opus 73.72 on the subset that ACE abstains (2{,}067 items) on. }
\label{tab:rerank}
\end{table}

\paragraph{Q2: the grammar as a candidate generator.} Here we focus on the examples ACE generates successfully on, to see how much the grammar helps in a hybrid system. We see that on this subset the top-10 often has the correct sentence but the ACE ranking is poor (Table~\ref{tab:rerank}): ACE ranks its realizations with the Redwoods parse-selection model (\texttt{redwoods.mem}, the only model its configuration loads) but it has no ranker trained for realization \citep{velldal2005,velldal2006}. Its first choice scores 61.82, the shortest candidate 69.01, Sonnet as a fluency judge 71.25 ($+9.4$, on this subset). The judge's pick also beats Sonnet's own generation by 4.2 points, and the hybrid scores 69.64 on the full split, four points over Sonnet alone. 

Adding Opus's own sentence to ACE's candidate pool raises the reference-selected score (corpus BLEU over the per-item best pick) to 83.07. The pooled judge (using Sonnet for fluency) reaches 78.36 on these items and 76.95 on the full split, essentially tied with the gold+silver system \citep{hajdik2019}. The judge (Sonnet) picks Opus's sentence on 66.9\% of items and an ACE candidate on the remaining 33.1\%, close to the 65.2\% of the reference selector, so the judge is not simply preferring model text.

\paragraph{Memorization} It does not by itself explain the realization results. A verbatim-continuation probe shows that Sonnet has seen some of the public text: it recalls 33\% of one essay's sentences and 0\% on three control profiles, but recall is unrelated to realization quality. A second control swaps one content noun for another in both the MRS and the reference (140 pairs). BLEU is 3.17 points higher on the original pairs than on the substituted ones, the direction memorization would predict, the LLMs are still on par with the trained systems.


\section{Parsing}
\label{sec:parse}

\begin{table}[t]
\centering\scriptsize\setlength{\tabcolsep}{2pt}
\begin{tabular}{@{}lrrrrrrrr@{}}
\toprule
 & \multicolumn{4}{c}{whole graph} & \multicolumn{2}{c}{relation F$_1$} & \multicolumn{2}{c}{recall} \\
\cmidrule(lr){2-5}\cmidrule(lr){6-7}\cmidrule(l){8-9}
 & EDM$_{\text{sf}}$ & EDM$_{\text{sa}}$ & exact & Smatch & lab. & unlab. & names & edges \\
\midrule
ACE 1214 & \textbf{91.0} & \textbf{88.3} & \textbf{50.5} & \textbf{94.3} & \textbf{90.5} & \textbf{91.1} & \textbf{92.8} & \textbf{94.8} \\
Opus 5 & 65.5 & 43.0 & 1.4 & 79.0 & 73.6 & 76.1 & 73.9 & 84.0 \\
Sonnet 4.5 & 57.2 & 36.9 & 0.6 & 77.6 & 66.5 & 69.7 & 65.5 & 74.1 \\
\midrule
\multicolumn{9}{@{}l}{\emph{Calibration inputs}} \\
gold & 100 & 100 & 100 & 100 & 100 & 100 & 100 & 100 \\
names junked & 0.0 & 79.4 & 0.0 & 79.2 & 99.7 & 99.6 & 0.0 & 0.0 \\
constant & 5.6 & 0.2 & 0.1 & 33.1 & 35.4 & 39.7 & 11.2 & 12.5 \\
\bottomrule
\end{tabular}
\caption{Parsing the same 998 sentences (text $\rightarrow$ SimpleMRS; LLMs three-shot, ACE top-1 parse). Outputs decode as SimpleMRS for 97.1\% (ACE, parse within resource limits), 96.3\% (Opus) and 99.4\% (Sonnet) of sentences; failures score 0. \emph{Whole graph}: EDM F$_1$ with spans off (sf) and span-anchored (sa); exact graph match ignoring spans; Smatch F$_1$ over all triples. \emph{Relation F$_1$}: Smatch on edges only, with role labels (lab.) or ignoring them (unlab., attachment only). \emph{Recall}: share of gold predicates named exactly (names), and share of gold edges recovered between correctly named nodes (edges). \emph{Calibration inputs} are scored against gold: gold itself, gold with every predicate name replaced by a nonsense string, and one fixed MRS for every sentence.}
\label{tab:parse}
\end{table}

\paragraph{Q3: the same models on the reverse task.} Both models write MRS that decodes, but rarely the gold MRS (Table~\ref{tab:parse}). ACE's top parse matches gold exactly on 50.5\% of sentences, but Opus does so on only 1.4\% and Sonnet on 0.6\%. On spans-off EDM the models trail ACE by 26 (Opus) and 34 (Sonnet) points.

The calibration rows show how to read these scores. Gold scores 100 on every metric. Gold with every predicate name junked (dummy word replacement) still scores 79.2 Smatch, level with Opus and above Sonnet, because Smatch rewards graph structure even when every name is wrong; a constant MRS still earns 33.1. Spans-off EDM and exact match fall to zero on both.

The failure is neither format, since 96--99\% of outputs decode, nor total since, counting only which predicate lemmas appear (\texttt{\_find\_v\_1} and \texttt{\_find\_v\_mental} both count as \emph{find}; F$_1$ over the multiset, ignoring edges), the models score 78 (Sonnet) and 84 (Opus). It lies in the grammar's conventions, at two levels. First, names: the models name only 65.48\% (Sonnet) and 74\% (Opus) of gold predicates exactly, against 93\% for ACE, and the misses are plausible but incorrect (\texttt{\_find\_v\_1} for \texttt{\_find\_v\_mental}, \texttt{compound} omitted). Second, attachment: between correctly named nodes they recover 74\% and 84\% of gold edges, against 95\% for ACE. These are wrong attachments, not wrong role labels. Ignoring labels raises relation F$_1$ only from 66.5 to 69.7 (Sonnet) and from 73.6 to 76.1 (Opus), far below ACE's 91.1.



\section{Conclusion}
We ran \citeauthor{hajdik2019}'s benchmark in both directions on the same sentences: from meaning representation to text, and from text back to meaning representation. The two directions give very different answers.

Going from MRS to text, prompted LLMs are strong. With three examples in the prompt, Sonnet matches the system trained on 72k gold pairs, and Opus beats it by about ten BLEU points and is level with the system that also used silver data. The sentences they write mostly keep the meaning of their input. Letting the models choose among the grammar's own candidate sentences helps Sonnet by four points and Opus slightly.

Going from text to MRS, the same models fail. They almost never produce the exact gold MRS. They mostly choose the right words but not the grammar's exact predicate names or the way it attaches arguments.

So a model can turn an MRS into a good sentence without being able to write that MRS itself. A high generation score should not be read as evidence that the model knows the representation.

\section*{Limitations}

We evaluate two Anthropic models under one prompting setup with three examples. Stronger setups, such as retrieval or constrained decoding, might narrow the parsing gap. Adding examples and enabling extended thinking did not close it. Our aim is to measure what prompted models can do on their own and combining them with constrained decoding is future work.

\bibliography{custom}

\appendix

\section{Appendix}
\label{app:spec}

\paragraph{Data.} \citet{hajdik2019}'s linearizer fails on 266 of the 10,201 test items (blank lines in their released source file). We score the remaining 9,935 minus the three held-out exemplars, 9,932 items. ACE generates for 7,865 of these.

\paragraph{Decoding.} Sonnet and the judge use temperature 0 and a 3,000-token limit. Opus~5 does not accept a temperature and runs at its default thinking.

\paragraph{Round-trip check.} ACE parses each system's sentence (top-1), and we score the parse against the input MRS with spans-off EDM. Sentences ACE cannot parse score zero. As a reference, the same procedure applied to the gold sentence gives 91.0 on the 1000 items.

\paragraph{Memorization controls.} The verbatim probe gives Sonnet the opening of a sentence from the test documents and checks whether it continues it word for word. The substitution control replaces one content noun with a different noun of the same length in both the MRS and the reference and compares BLEU on original and substituted pairs.

\paragraph{Parsing metrics.} We read gold and predicted SimpleMRS with PyDelphin~1.11 and convert both to EDS graphs. An output that cannot be read scores zero on every metric.

\emph{EDM} splits a graph into small facts: which predicates it contains, which arguments link them, and which features (tense, number, and so on) they carry. It then counts how many of the gold facts the prediction also contains, reporting precision, recall and F$_1$. In the \emph{span-anchored} version, a predicted node matches a gold node only if both cover the same stretch of the sentence. In the \emph{spans-off} version, nodes match by predicate name alone.

\emph{Exact match} asks whether the predicted graph is the gold graph with only the variable labels changed, ignoring spans. We check this with a standard graph-matching search, which finished on every item.

\emph{Smatch} is the usual AMR metric, applied to the EDS graphs. It finds the pairing of predicted and gold nodes that makes the most facts agree, then scores those facts. It searches for this pairing at random, so repeated runs can differ by about 0.3.

\emph{Predicate-lemma overlap} ignores edges and sense labels. \texttt{\_find\_v\_1} and \texttt{\_find\_v\_mental} both count as \emph{find}, and we compute F$_1$ between the two lists of lemmas. \emph{Edge recall} pairs each gold node with a predicted node that has exactly the same predicate name. It then asks what share of the gold edges between paired nodes also appear, with the same label, in the prediction.

\paragraph{BLEU.} We use SacreBLEU~2.6 with its default settings (13a tokenization, exponential smoothing, one reference). 

\paragraph{Choosing among candidates.} The \emph{reference selector} picks, for each item, the candidate closest to the reference sentence by BLEU. It looks at the answer, so it only shows how good the candidate list could be; no real system can use it. Picking the best sentence one at a time does not quite maximize corpus BLEU, so we also report a \emph{corpus-BLEU selector}. It starts from the same picks and keeps swapping candidates as long as corpus BLEU goes up.

\paragraph{Prompts.} Each prompt has a system message, then three worked examples, then the test item.
\begin{itemize}\itemsep0pt
\item \emph{Realization.} ``You are an expert in the English Resource Grammar (ERG) and Minimal Recursion Semantics (MRS). You are given a linearized Dependency MRS graph. Write the English sentence it represents. Tokens like named0, named1, month0 are placeholders for named entities: copy them through verbatim in your output. Output only the sentence, with no explanation, no quotes and no markup.'' Examples are \texttt{Graph:}/\texttt{Sentence:} pairs.
\item \emph{Parsing.} ``You are an expert linguistic annotator producing Minimal Recursion Semantics in the DELPH-IN SimpleMRS format used by the English Resource Grammar. Output ONLY the MRS.'' Examples are \texttt{Sentence:}/\texttt{MRS:} pairs.
\item \emph{Judge.} ``You are a careful editor of written English. You will be shown numbered candidate sentences that are all intended to express the same meaning. Choose the single most fluent, natural and well-formed English sentence. Output only its number.'' The user message lists the candidates and ends with ``Answer with the number only.''
\end{itemize}

\end{document}